**Tittle:**

A comment on the paper "Prediction of Kidney Function from Biopsy Images using Convolutional Neural Networks"

**Authors:**

Washington LC dos-Santos[1], Angelo A Duarte[2], Luiz AR de Freitas[1]

[1] Fundação Oswaldo Cruz, Instituto Gonçalo Moniz, Salvador, BA, Brazil.

[2] Universidade Federal de Estadual de Feira de Santana, Laboratório de Computação de Alto Desempenho, Feira de Santana, BA, Brazil.

**Category:**

LG - Machine Learning (stat.ML); Quantitative Methods (q-bio.QM)

**Letter**

Dear Sir/Madam,

The paper "Prediction of Kidney Function from Biopsy Images using Convolutional Neural Networks" by Ledbetter et al. (2017)(Ledbetter et al., 2017) deals with the interesting subject of automatized estimate of kidney function from histological images. The authors report that using Convolutional Neural Network (CNN) associated with some image processing techniques of renal biopsies the 12 mo glomerular filtration rate can be predicted with a mean absolute error of 17.55 ml/min. Notwithstanding the importance of the work and the huge effort made for preparing the dataset and for configuring and training the CNN, there are some points that require attention and a prompt revision of the published article and the of the presented data. The figures 2, 5 and 6 from the Ledbetter et al. (2017) paper represent skin sections and not kidney samples. The skin sections in figure 2, 5 and 6 show an intradermal nevus and the structure showed in detail is, in fact a dilated hair follicle containing a keratin plug. Misusing skin instead of kidney sections for training or validating the system may compromise the accuracy of the proposed estimate. Histological sections of normal skin (a and b) and normal kidney (c and d), with glomerulus in detail (d) are shown in the figure accompanying this letter for effect of comparison. Moreover, information on the amount of renal cortical tissue required for the estimates of glomerular filtration rate is missing. Renal biopsies containing at least 8 glomeruli are usually accepted as adequate renal cortex representation for most diagnosis in Nephropathology (Corwin et al., 1988; Fogo, 2003). Finally, we neither found the number of images used in the training dataset nor the amount of images used to validation dataset. According to the article, it seems that images from 60 patients were used to create the training dataset, however, it is not clear how many images were generated from these patients' biopsies.

**References:**

Corwin, H.L., Schwartz, M.M., Lewis, E.J., 1988, The importance of sample size in the interpretation of the renal biopsy. Am J Nephrol 8, 85-89.

Fogo, A.B., 2003, Approach to renal biopsy. Am J Kidney Dis 42, 826-836.

Ledbetter, D., Ho, L., Lemley, K.V., 2017, Prediction of Kidney Function from Biopsy Images Using Convolutional Neural Networks. arXiv preprint arXiv:1702.01816.

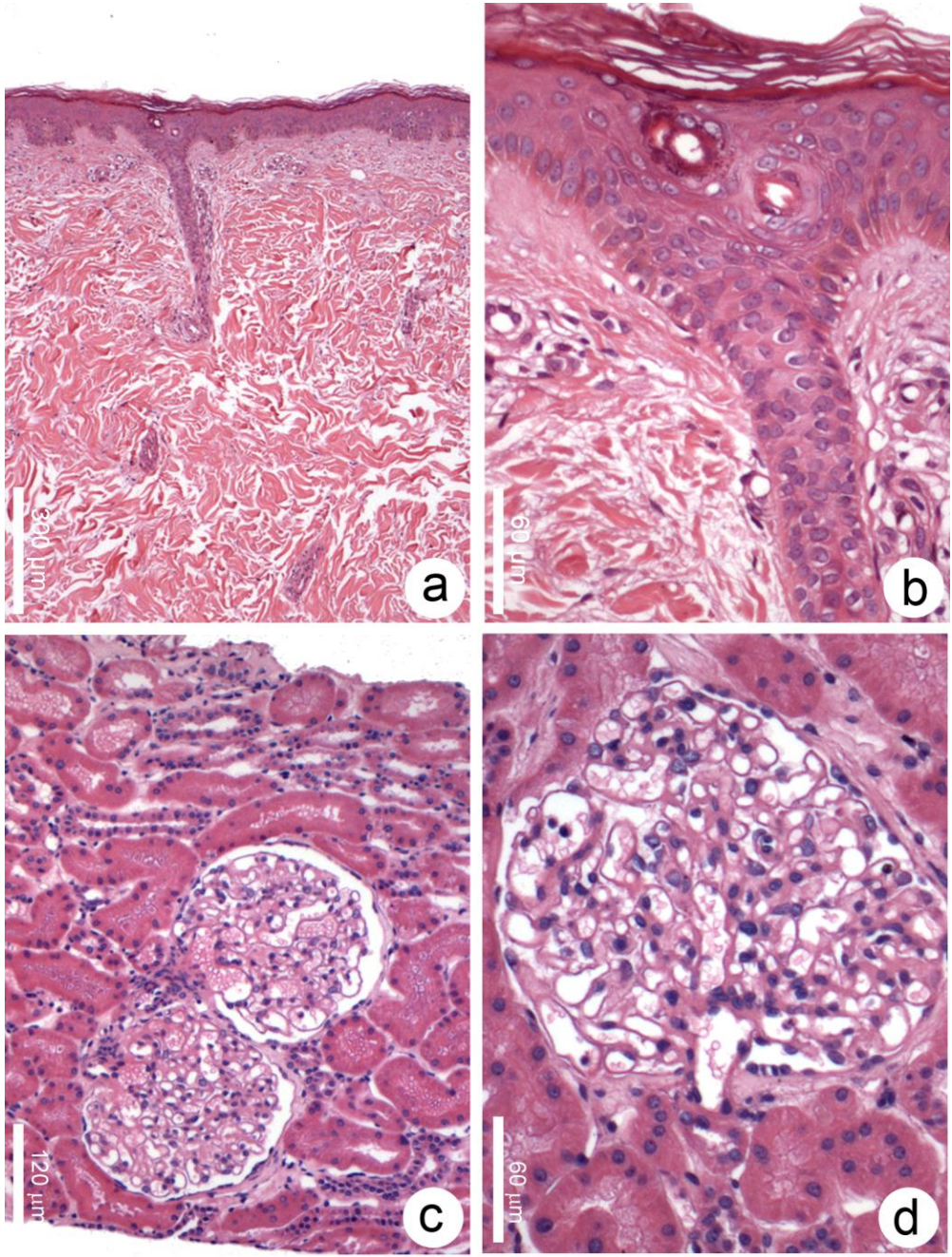